\documentclass{article}

\usepackage{microtype}

\usepackage[nolist]{acronym}

\usepackage{enumitem}

\usepackage{amsmath}
\usepackage{bbm}
\usepackage{mathtools}
\usepackage{braket} %
\usepackage{siunitx}
\usepackage{xspace}
\usepackage{balance}

\usepackage{textcomp}

\usepackage[]{subcaption}

\usepackage{booktabs}
\usepackage{multirow}
\usepackage{tabularx}

\newcolumntype{C}[1]{>{\centering\arraybackslash}p{#1}}

\usepackage{pifont} %
\usepackage{wasysym} %

\usepackage{pgfplots} %
\usepackage{tikz}
\usepackage{forest}

\usepackage{svg}

\pgfplotsset{compat=1.13}
\usetikzlibrary{calc}
\usetikzlibrary{shadows}
\usetikzlibrary{positioning,arrows.meta,fit,matrix}
\usetikzlibrary{backgrounds}
\usetikzlibrary{patterns}
\usetikzlibrary{math}
\usetikzlibrary{shapes}
\usetikzlibrary{perspective}
\usetikzlibrary{shapes.arrows} %
\usetikzlibrary{chains}
\usepgfplotslibrary{groupplots}
\usepgfplotslibrary{statistics}
\usepgfplotslibrary{fillbetween}

\definecolor{cbone}  {HTML}{006BA4} %
\colorlet{primarycolor}{cbone}

\tikzset{
	font=\footnotesize,
	container/.style={draw, rectangle, rounded corners, dashed, inner
		sep=0em, darkgray, thick},
	connectuni/.style={thick, shorten >= 0.1em, shorten <= 0.1em},
	connectdir/.style={->,>=stealth',connectuni},
	vectorconnect/.style={thick, ->, dotted, shorten >= 0.1em, shorten 
	<= 0.1em}, %
}
\pgfplotsset{
    legend image code/.code={
        \draw[mark repeat=2,mark phase=2]
        plot coordinates {
            (0cm,0cm)
            (0.15cm,0cm)  %
            (0.3cm,0cm)   %
        };%
    }
}

\usepackage{dirtytalk}

\usepackage[most]{tcolorbox}

\usepackage{xcolor}
\usepackage{colortbl}
\usepackage{url}
\usepackage{relsize}

\usepackage{graphicx}
\usepackage{xparse}

\usepackage{mathtools}

\usepackage{caption}

\usepackage{array}
\usepackage{makecell}
\usepackage{multirow}
\usepackage{rotating}

\usepackage{tablefootnote}

\usepackage{tcolorbox}
\usepackage{blindtext}

\usepackage[linesnumbered,ruled,vlined]{algorithm2e}

\usepackage{ifthen}

\usepackage[short]{optidef}
\usepackage{hyperref}
\usepackage{numprint}
\npstyleenglish

\vfuzz=\hfuzz

\definecolor{cb-black}      {RGB}{  0,   0,   0}
\definecolor{cb-blue-green} {RGB}{  0,  073,  073}
\definecolor{cb-green-sea}  {RGB}{  0, 146, 146}
\definecolor{cb-rose}       {RGB}{255, 109, 182}
\definecolor{cb-salmon-pink}{RGB}{255, 182, 119}
\definecolor{cb-purple}     {RGB}{ 73,   0, 146}
\definecolor{cb-blue}       {RGB}{ 0, 109, 219}
\definecolor{cb-lilac}      {RGB}{182, 109, 255}
\definecolor{cb-blue-sky}   {RGB}{109, 182, 255}
\definecolor{cb-blue-light} {RGB}{182, 219, 255}
\definecolor{cb-burgundy}   {RGB}{146,   0,   0}
\definecolor{cb-brown}      {RGB}{146,  73,   0}
\definecolor{cb-clay}       {RGB}{219, 209,   0}
\definecolor{cb-green-lime} {RGB}{ 36, 255,  36}
\definecolor{cb-yellow}     {RGB}{255, 255, 109}

\newtcbtheorem{Summary}{\bfseries Summary}{enhanced,drop shadow={black!50!white},
  coltitle=black,
  top=0.3in,
  attach boxed title to top left=
    {xshift=1.5em,yshift=-\tcboxedtitleheight/2},
  boxed title style={size=small,colback=pink}
}{summary}

\definecolor{colpositive}{RGB}{11, 49, 66}
\definecolor{colnegative}{RGB}{170,72,72}

\definecolor{tikpositive}{RGB}{11, 49, 66}
\definecolor{tiknegative}{RGB}{170,72,72}

\definecolor{myblue}{RGB}{11, 49, 66}
\definecolor{myred}{RGB}{170,72,72}

\definecolor{intelblue}{HTML}{7AAFDA}
\definecolor{nvidiagreen}{HTML}{8CC984}
\definecolor{blizOrange}{HTML}{E9AB4C}
\definecolor{appleGray}{HTML}{C0C0C0}

\definecolor{colorVit}{RGB}{128, 180, 230}
\definecolor{colorResnet}{RGB}{252, 90, 40}
\definecolor{colorEfficientnet}{HTML}{4D805F}

\DeclareMathOperator*{\argmax}{arg\,max}
\DeclareMathOperator*{\argmin}{arg\,min}

\newcommand{\xfool}{\hat{x}}
\newcommand{\yfoollabel}{t}

\newcommand{\logitfunction}{f}
\newcommand{\predictionfunction}{F}
\newcommand{\originalmodelparameters}{\bar{\theta}}
\newcommand{\modelparameters}{\theta}

\newcommand{\hardware}{h}
\newcommand{\hardwareone}{h_1}
\newcommand{\hardwaretwo}{h_2}
\newcommand{\hardwareswap}[1]{\mkern-8mu\xrightarrow{\raisebox{-1.5pt}[0pt][0pt]{\ensuremath{\scriptstyle{#1}}}}\mkern-5mu}

\newcommand{\loss}{\mathcal{L}}

\newcommand{\weightmatrix}{W}
\newcommand{\permutedweightmatrix}{\hat{\weightmatrix}}

\usepackage{hyperref}

\usepackage[preprint]{icml2026}

\usepackage{amsmath}
\usepackage{amssymb}
\usepackage{mathtools}
\usepackage{amsthm}
\usepackage{balance}

\usepackage{xcolor}

\definecolor{fcolor}{HTML}{CB532F}
\definecolor{pcolor}{HTML}{2f6676}
\definecolor{defaultcolor}{RGB}{200, 200, 200}
\definecolor{dcolor}{RGB}{160, 180, 120}
\definecolor{ngreen}{HTML}{2f8556}

\usepackage[capitalize,noabbrev,nameinlink]{cleveref}

\theoremstyle{plain}

\theoremstyle{definition}

\theoremstyle{remark}

\usepackage[textsize=tiny]{todonotes}

\icmltitlerunning{Hardware-Triggered Backdoors}



\newcommand{\code}[1]{\textcolor{red!50!black}{#1}}

\begin{document}

\twocolumn[
  \icmltitle{Hardware-Triggered Backdoors}

  \icmlsetsymbol{equal}{*}

  \begin{icmlauthorlist}
    \icmlauthor{Jonas Möller}{bifold,tuberlin}
    \icmlauthor{Erik Imgrund}{bifold,tuberlin}
    \icmlauthor{Thorsten Eisenhofer}{cispa}
    \icmlauthor{Konrad Rieck}{bifold,tuberlin}
  \end{icmlauthorlist}

  \icmlaffiliation{bifold}{Berlin Institute for the Foundations of Learning and Data
(BIFOLD), Germany}
  \icmlaffiliation{tuberlin}{TU Berlin, Germany}
  \icmlaffiliation{cispa}{CISPA Helmholtz Center for Information Security, Germany}

  \icmlcorrespondingauthor{Jonas Möller}{jonas.moeller.1@tu-berlin.de}

  \icmlkeywords{Machine Learning, ICML}

  \vskip 0.3in
]

\printAffiliationsAndNotice{}  %

\begin{abstract}
    Machine learning models are routinely deployed on a wide range of computing hardware. Although such hardware is typically expected to produce identical results, differences in its design can lead to small numerical variations during inference.
    In this work, we show that these variations can be exploited to create backdoors in machine learning models. The core idea is to shape the model's decision function such that it yields different predictions for the same input when executed on different hardware. This effect is achieved by locally moving the decision boundary close to a target input and then refining numerical deviations to flip the prediction on selected hardware.
    We empirically demonstrate that these hardware-triggered backdoors can be created reliably across common GPU accelerators. Our findings reveal a novel attack vector affecting the use of third-party models, and we investigate different defenses to counter this threat.
\end{abstract}

\section{Introduction}

Hardware acceleration is a cornerstone of machine \mbox{learning} inference. Depending on the application, learning models are routinely deployed on a wide range of computing hardware, from inexpensive consumer GPUs to high-performance accelerators. While these devices differ significantly in efficiency and energy consumption, a key assumption is that they compute identical results, enabling seamless deployment across heterogeneous setups. 
Interestingly, this assumption does not fully hold in practice. Differences in hardware design and floating-point behavior give rise to small numerical variations when the same model is executed on different devices~\citep{schlogl2024causes}. These deviations can complicate the comparison of model outputs but are generally considered harmless.

In this paper, we show that minor deviations induced by hardware can be far from harmless. Following recent work on numerical variations during inference~\citep{ZhaFoeMul+24, moller2025adversarial, yuan2025understanding}, we make an unsettling observation: in reality, a trained model does not correspond to a single decision function; instead, it gives rise to a family of highly similar yet distinct functions, depending on the employed hardware. While these functions remain numerically close to each other in benign settings, an adversary may attempt to target their gap to activate malicious behavior on selected hardware. We refer to this novel attack type as a \emph{hardware-triggered backdoor}.

\begin{figure}[t]
\centering
\includegraphics[width=0.40\textwidth]{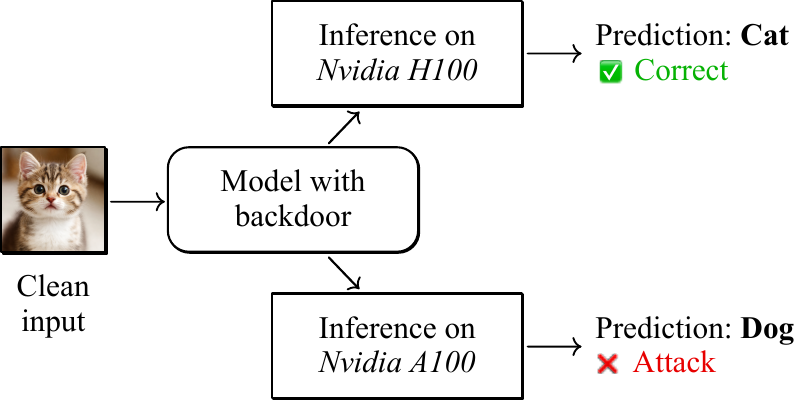}
\caption{Attack overview: The backdoor is triggered depending on the hardware accelerator used for inference.}
\label{fig:overview}
\end{figure}

To explore the feasibility of this attack, we introduce a method for manipulating a model’s decision function so that it yields conflicting predictions for a selected input when executed on different hardware. We achieve this effect by locally moving the decision boundary close to the input and then amplifying numerical deviations to flip the prediction on selected hardware. Unlike traditional attacks, this backdoor does not employ an explicit trigger in the input. The numerical behavior of the hardware acts as a latent trigger for flipping the prediction. \cref{fig:overview} illustrates the general working principle of this attack type.

We empirically find that this approach is effective and independent of accidental numerical instability. Instead, hardware-dependent deviations can be induced in a controlled manner for common model architectures, enabling attack success rates above 90\% with no impact on model performance. The resulting backdoors can be targeted to particular hardware accelerators and made robust to non-trivial changes in inference, such as input perturbations, batching, and mixed-precision inference.

We conclude that hardware-triggered backdoors pose a threat whenever third-party models are deployed across device setups. As a countermeasure, we investigate different defenses and evaluate their effectiveness, with positive results. Our findings highlight that the security of machine learning must be considered from  trained models down to the underlying hardware, as numerical deviations, even if seemingly small, may have adversarial effects.

In summary, we make the following major contributions:
\begin{enumerate}[leftmargin=*]
    \item \textbf{Hardware-triggered backdoors.} We introduce a new type of backdoor in which malicious behavior is activated by specific hardware rather than an input trigger.     
    \item \textbf{Causal analysis of differences.} We perform a layer-wise causal analysis to identify where hardware-induced differences arise during inference.
    \item \textbf{Evaluation of efficacy.} We demonstrate the efficacy of hardware-triggered backdoors over different model architectures, hardware devices, and input perturbations.
\end{enumerate}

\section{Numerical Deviations}
\label{sec:background}

In theory, inference in machine learning models is a well-defined process in which an input is passed through a decision function using learned parameters. From this perspective, operations such as matrix multiplication or convolution are precisely specified, leaving no room for  deviations.
In practice, however, these operations are performed using floating-point numbers of limited precision. While this precision can be adapted, it remains finite, rendering arithmetic inherently imprecise~\citep{IEEE754-2019}.

\paragraph{Non-associativity.}
The primary source of this imprecision is the non-associativity of floating-point addition, where we can have $a + (b + c) \neq (a + b) + c$ \citep{schlogl2024causes}. 
That is, the result of a sum also depends on the order in which terms are added. Many operations, including matrix multiplication, convolution, aggregations, and attention, rely on a series of additions. The order of these additions is shaped by the underlying hardware resources, for instance through choices of block size and warp scheduling.

We can illustrate this effect by considering a simple matrix $M \in \mathbb{R}^{100\times100}$ whose entries are all equal to $0.01$. When computing the squared Frobenius norm of $M$ on two GPUs using the source code given in \cref{lst:torch-trace} in the appendix, we observe slightly deviating results,
\begin{align*}
||M||_F^2 & = \operatorname{tr}(M^T M) = 1\\ 
 & \approx 0.9999999403953552 ~~\text{(Nvidia A100)} \\
 & \approx 0.9999990463256836 ~~\text{(Nvidia H100)}.
\end{align*}

As the two GPUs have distinct hardware features~\citep{nvidiaA100, nvidiaH100}, different kernel implementations are selected for the matrix multiplication, each suitable for the specific device. Consequently, intermediate results are grouped differently on the two GPUs, changing the order in which summands are combined and thereby leading to slight deviations of the squared Frobenius norm.

\paragraph{Notation.}
To formalize these differences, we introduce the following notations. We consider a learning model $\modelparameters$ that induces a theoretical decision function
$\logitfunction_{\modelparameters} \colon \mathbb{R}^n \rightarrow \mathbb{R}^c$,
which maps an $n$-dimensional input to $c$ class logits. In practice, the realized behavior of this function depends on the hardware $\hardware \in \mathbb{H}$ on which the model is executed, where $\mathbb{H}$ denotes a set of functionally equivalent hardware platforms, such as different GPU accelerators. Accordingly, we model the effective decision function as
\begin{equation}
    \logitfunction_{\modelparameters} \colon \mathbb{R}^n \times \mathbb{H} \longrightarrow \mathbb{R}^c,
\end{equation}
thereby making explicit its joint dependence on the input and the deployed computing hardware.

This hardware dependence can induce unexpected discrepancies during inference. When predictions are obtained by selecting the class with the largest logit, that is,
\begin{equation}
\predictionfunction_{\modelparameters}(x;\hardware)
= \arg\max_i \logitfunction_{\modelparameters}(x;\hardware)_i,
\end{equation}
where $f(\cdot)_i$ denotes the $i$-th logit, it may occur that
\begin{equation}
\predictionfunction_{\modelparameters}(x;\hardwareone)
\neq
\predictionfunction_{\modelparameters}(x;\hardwaretwo),
\end{equation}
for an input $x$ and two devices $\hardwareone$ and $\hardwaretwo$. In such cases, identical inputs processed by the same model are assigned different class labels solely due to the underlying hardware.

\section{Hardware-Triggered Backdoors}
\label{sec:backdoors}

Fortunately, numerical deviations arise stochastically during inference and, due to their small magnitude, rarely influence class predictions. Hence, they are generally regarded as harmless. We challenge this view by exploring whether hardware-based deviations can be deliberately exploited to create backdoors in learning models.

\subsection{Threat Model}

In our threat model, the victim employs different hardware devices for inference of a learning model. For example, smaller devices may be used during development, while more powerful accelerators are used in production systems. We assume that the attacker is aware of this heterogeneity and can manipulate the model prior to deployment, for instance through a supply-chain attack, by acting as the model provider, or by uploading a manipulated model to a public platform, such as HuggingFace.

The attacker’s goals are twofold. First, they aim to trigger misclassification of selected inputs only on specific devices of the victim, such as those used in production. This allows the attack to remain stealthy even if the target inputs are inspected on development devices for forensic analysis. Second, they seek to preserve the original model's behavior on all other inputs, ensuring that the manipulation remains stealthy during regular operation. %

\subsection{Attack Strategy}
\label{sec:backdoor-construction}

Let us begin by considering a target input $\xfool$ that the adversary aims to misclassify on one hardware device $\hardwareone$ but not on another device $\hardwaretwo$. The attacker’s objective is to induce a deviation such that
\begin{equation}
\label{eq:f_not_equal_f}
\predictionfunction_{\modelparameters}(\xfool;\hardwareone)
\neq
\predictionfunction_{\modelparameters}(\xfool;\hardwaretwo),
\end{equation}
while ensuring that, for all other inputs, the behavior of the modified model $\modelparameters$ remains indistinguishable from that of the original one. Note that in this case $\xfool$ is deliberately selected, whereas in common errors caused by numerical deviations the affected inputs arise at random.

At a first glance, implementing this attack appears straightforward: one could formulate a loss that enforces conflicting predictions across hardware and optimize it using gradient-based methods, similar to existing backdoor attacks~\citep{liu2018trojaning,shafahi2018PoisonFrogs}. Unfortunately, this approach is infeasible. First, numerical deviations are hardware-specific, and so are the corresponding gradients. Second, the deviations are non-differentiable, ruling out the tools commonly used in backdoor attacks.

To overcome these challenges, we build on a key observation: in practice, a learning model does not induce a single decision function, but rather a family of closely related functions that depend on the underlying hardware. While these functions are numerically close, their differences can become consequential in regions where the model is sensitive to small perturbations. In particular, when an input lies close to a decision boundary, even minor deviations may suffice to change the predicted label.

\begin{figure}[h]
  \centering
  \begin{subfigure}[t]{0.45\textwidth}
    \centering
    \includegraphics[width=0.7\textwidth]{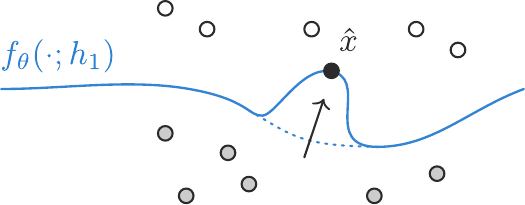}
    \vspace{5pt}
    \caption{Step 1: Shaping of decision boundary}
    \label{fig:steps-step2}
  \end{subfigure}
  \begin{subfigure}[t]{0.45\textwidth}
    \centering
    \vspace{15pt}
    \includegraphics[width=0.7\textwidth]{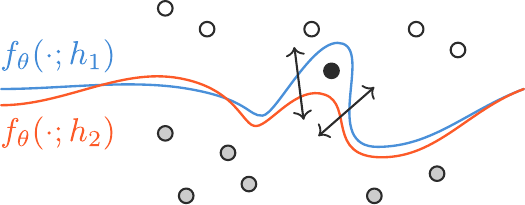}
    \vspace{5pt}
    \caption{Step 2: Refinement of deviation}
    \label{fig:steps-step3}
  \end{subfigure}
  \caption{Construction of hardware-triggered backdoors: (a) The decision boundary is moved close to the target $\xfool$; (b) Hardware deviations are amplified between $\hardwareone$ and $\hardwaretwo$.}
  \label{fig:steps}
\end{figure}

Based on this observation, we propose a two-step attack strategy, as illustrated in \cref{fig:steps}. First, we locally move the decision boundary into the vicinity of a target input on one hardware device (\cref{fig:steps-step2}). Second, we adjust the numerical deviations between two devices, such that the respective decision functions disagree on the prediction of the input (\cref{fig:steps-step3}).

\subsection{Shaping the Decision Boundary}

For the first step, we treat the model as a single decision function and operate on it using standard gradient-based optimization. To this end, we optimize a proxy loss $\loss$ that encourages proximity to the decision boundary while constraining deviations from the original model behavior. All computations are performed on a single hardware device~$\hardwareone$. The resulting optimization problem is
\begin{equation}
\label{eq:attack-formulation}
    \argmin_{\modelparameters} \; \loss\!\left(\modelparameters, \logitfunction_{\modelparameters}(\xfool; \hardwareone)\right) \, 
\end{equation}
where the proxy loss $\loss$ consists of three components,
\begin{align}
    \loss(\modelparameters,y)
    &= \alpha \loss_{\text{diff}}(\modelparameters,y)
    + \beta \loss_{\text{class}}(\modelparameters,y)
    + \gamma \loss_{\text{reg}}(\modelparameters)\,.
    \label{eq:proxy-loss}
\end{align}
and $\alpha$, $\beta$, and $\gamma$ control the relative influence of the individual terms during optimization.

The first loss term encourages proximity of $\xfool$ to the decision boundary by minimizing the difference between the two largest logits, effectively creating a tie,
\begin{equation}
    \loss_{\text{diff}}(\modelparameters, y)
    = \max_i y_i - \max_{j \neq \argmax_i y_i} y_j \, .
    \label{eq:first-term}
\end{equation}
The second term penalizes deviations from the original source class $\yfoollabel$ of $\xfool$,
\begin{equation}
    \loss_{\text{class}}(\modelparameters, y)
    = \max\,\bigl( \max_{i \neq \yfoollabel} y_i - y_{\yfoollabel}, 0 \bigr) \, .
    \label{eq:second-term}
\end{equation}
This ensures that the input sample stays close to the original label and so the backdoor remains stealthy on one device.
Finally, to keep the manipulation localized, we regularize deviations from the original unmodified model $\originalmodelparameters$,
\begin{equation}
    \loss_{\text{reg}}(\modelparameters)
    = \lVert \modelparameters - \originalmodelparameters \rVert^2 \, .
    \label{eq:third-term}
\end{equation}

For differentiable models, \cref{eq:attack-formulation,eq:proxy-loss,eq:first-term,eq:second-term,eq:third-term} can be optimized directly using standard gradient descent. %

\subsection{Refining Deviations}

The target input $\xfool$ now lies in the immediate vicinity of the decision boundary, yet it is still likely to receive the same prediction across all hardware platforms. The goal of the second step is to exploit this fragile configuration and amplify hardware-dependent divergence. As these deviations are non-differentiable, however, we must resort to heuristic strategies to manipulate them across hardware devices.
In particular, we consider two types of manipulations:
\begin{itemize}
    \item \textit{Implicit modifications.}
    This type preserves mathematical equivalence under exact arithmetic but alters the order of floating-point operations, resulting in platform-dependent deviations.
    
    \item \textit{Explicit modifications.}
    This type slightly changes model parameters and therefore affects the computation even under exact arithmetic. While this mechanism is more powerful, it may affect model utility.
\end{itemize}

Different realizations of these strategies are conceivable, including reformulating operators, altering numerical representations, or introducing low-level manipulations. For simplicity, we focus on one representative strategy for each type and leave a more exhaustive analysis to future work. The ablation study in \cref{sec:ablation} demonstrates the effectiveness of both strategies.

\paragraph{Implicit modification: Topological permutation}
As an instance of implicit modifications, we introduce a \emph{topological permutation}, which alters the order of additions. Specifically, we use a permutation matrix $P_i$ and its inverse $P_i^{-1}$ to permute weights of the model. Given a matrix multiplication $\weightmatrix_1 \weightmatrix_2$ inside the model, we construct:
\begin{align}
    \weightmatrix_1 \weightmatrix_2 
    &= \underbrace{(P_1\weightmatrix_1)}_{\permutedweightmatrix_1}\underbrace{(P^{-1}_1\weightmatrix_2)}_{\permutedweightmatrix_2}\, .
\end{align}
This construction applies to models with at least two consecutive linear layers, a pattern present in different architectures, including transformers. %

Depending on the choice of $P$, this strategy yields different realizations of the same multiplication, whose computation differs only through numerical deviations arising from the permuted parameter topology.

\paragraph{Explicit modification: Parameter perturbation.}  As an explicit modification, we consider small perturbations of the model parameters themselves. Concretely, we select a set of $k$ bits in the parameters and flip their values, thereby introducing limited numerical changes into the computation. Unlike implicit modifications, such perturbations alter parameter values and thus affect the computation even under exact arithmetic. 

From a methodological view, this types of modifications provides a more direct means of modifying hardware-dependent effects, while keeping the overall modification constrained to $k$ bits.

\subsection{Alternating Optimization}

Finally, we combine both steps in an alternating optimization procedure. In each iteration, the decision boundary is first locally shifted toward the target input. Subsequently, both strategies are applied to search for a split decision across the selected hardware. To address the heuristic nature of the manipulation strategies, we construct $m$ candidate models in each iteration. We terminate once a model exhibits a functional backdoor. Moreover, during optimization we discard  candidates that fail to preserve a selected level $\rho$ of the original model's performance.

\section{Evaluation}
\label{sec:main-attack}

Equipped with an approach for exploiting numerical deviations, we are ready to 
empirically investigate hardware-triggered backdoors. First, we assess their efficacy for a single target input and a pair of hardware devices. We then generalize the setup to multiple target inputs and groups of devices. Finally, we present an ablation study of our approach. To foster reproducibility, we release the source code of our experiments at \url{https://github.com/mlsec-group/hardware-triggered-backdoors}.

\begin{table}[h]
    \centering \small
    \caption{Overview of considered GPU platforms.}
    \begin{tabular}{lll}
    \toprule
    \textbf{GPU} & \makecell[c]{\textbf{Architecture}} & \makecell[c]{\textbf{Chip}} %
    \\
    \midrule
    Nvidia H100 & Hopper & GH100
    \\
    Nvidia A100  & Ampere & GA100
    \\
    Nvidia A100 (MIG-40GB) & Ampere & GA100
    \\
    Nvidia A40 & Ampere & GA102
    \\
    Nvidia Quadro RTX 6000 & Turing & TU102
    \\
    \bottomrule
    \end{tabular}
    \label{tab:set-of-gpus}
\end{table}

\paragraph{Hardware platforms.} 
We consider five common Nvidia GPUs, spanning four architectural generations (Table~\ref{tab:set-of-gpus}). These devices differ in microarchitectural details and supported numerical formats, making them well suited for studying hardware-dependent behavior. 
Moreover, we consider \code{float32}, \code{float16}, and \code{bfloat16} as widely used numerical formats on these devices. %

A particularly challenging case in our setup is the A100 (MIG-40GB). While it is identical in hardware to the standard A100, the use of Multi-Instance GPU (MIG) introduces a virtualization layer that slightly alters the execution environment. As we show later, this difference can be sufficient to trigger backdoors in some models. Other models, however, remain unaffected and exhibit bit-identical behavior across these two devices.

We restrict our experiments to devices from a single manufacturer. In this setting, differences in computation can originate only from the employed hardware devices, whereas experiments across manufacturers would also induce numerical deviations due to their different software backends~\citep{moller2025adversarial}. As these software-induced deviations would further enlarge the attack surface, we consider our setup a conservative basis for evaluating hardware-triggered backdoors.

\paragraph{Models and data.} 
As hosts for the backdoors, we consider three common vision models. Specifically, we employ \emph{ResNet-18}~\cite{he2016deep} and \emph{EfficientNetV2-S}~\cite{tan2021efficientnetv2}, which primarily rely on convolutional layers, as well as a \emph{Vision Transformer (ViT)} with a $32 \times 32$ input patch size~\cite{DBLP:conf/iclr/DosovitskiyB0WZ21}.
All models are initialized from public pretrained weights, and all experiments are conducted on ImageNet~\cite{deng2009imagenet}. Target inputs are sampled uniformly at random from the training set so as not to affect clean performance.

\paragraph{Attack setup.}
We follow the two-step attack procedure described in~\cref{sec:backdoors}. After a preliminary study, we fix $\beta = 0.1$ and $\gamma = 10{,}000$, and use an adaptive schedule for $\alpha$ to gradually adjust the influence of the decision-boundary term with $500$ gradient descent steps per iteration. Moreover, we set the number of bit flips to $k = 5$ and the number of candidate models to $m = 256$, with 128 candidates refined with permutations and 128 with bit flips. We define $\rho = 95\%$ as the minimum performance that must be retained. 
In this configuration, the attack already yields satisfactory results after six iterations, and we therefore use it throughout all experiments. Finally, all experiments use a batch size of one to avoid numerical deviations induced by batching.

\subsection{Attack Efficacy}
\label{sec:attack-eval}

As a first experiment, we investigate the efficacy of our backdoor across pairwise combinations of hardware platforms. In particular, for each pair of devices, we conduct the attack 100 times using independently sampled target inputs. Each run starts from a fresh copy of the pretrained model and applies up to six iterations of the two-stage approach described in Section~\ref{sec:backdoors}. We repeat this experiment for \code{float32}, \code{float16}, and \code{bfloat16} as model data types.

\begin{table}[b]
\small
    \centering
    \caption{Attack success rate for models in \code{float32}.}
\begin{tabular}{lccc}
\toprule
\textbf{GPU} & \textbf{ViT} & \textbf{ResNet} & \textbf{EfficientNet} \\
\midrule
H100        & $94\thinspace\% \pm 6\thinspace\%$ & $100\thinspace\% \pm 0\thinspace\%$ & $100\thinspace\% \pm 0\thinspace\%$ \\
A100        & $98\thinspace\% \pm 3\thinspace\%$ & $75\thinspace\% \pm 43\thinspace\%$ & $100\thinspace\% \pm 0\thinspace\%$ \\
A100-MIG40  & $98\thinspace\% \pm 3\thinspace\%$ & $75\thinspace\% \pm 43\thinspace\%$ & $100\thinspace\% \pm 0\thinspace\%$ \\
A40         & $94\thinspace\% \pm 6\thinspace\%$ & $100\thinspace\% \pm 0\thinspace\%$ & $100\thinspace\% \pm 0\thinspace\%$ \\
RTX6000     & $94\thinspace\% \pm 6\thinspace\%$ & $100\thinspace\% \pm 0\thinspace\%$ & $100\thinspace\% \pm 0\thinspace\%$ \\
\bottomrule
\end{tabular}
    \label{tab:main-float32}
\end{table}

\paragraph{Results.}
The results for \code{float32} are summarized in Table~\ref{tab:main-float32}, while the corresponding measurements for \code{float16} and \code{bfloat16} are reported in Tables~\ref{tab:main-float16} and~\ref{tab:main-bfloat16} in the appendix.

We find that hardware-triggered backdoors can be created reliably across almost all evaluated models, GPUs and data types. The sole exception is the A100 and A100-MIG40 pair on ResNet. In this case, the two devices exhibit bit-identical behavior, as the virtualization does not affect model behavior. For all other pairs, we attain attack success rates above $94\%$.
Furthermore, the backdoored models retain a median of 99.8\% of the original model performance, demonstrating the efficacy of the attack.

Interestingly, the backdoors created by our approach are even highly effective under full-precision data types (\code{float32}), despite prior work suggesting increased numerical precision as a potential mitigation against hardware-induced effects~\citep{yuan2025understanding}.

\subsection{Multiple Target Inputs}
\label{sec:eval-multi}

Thus far, we have focused on creating backdoors for a single target input. In practice, however, an attacker may wish to implant multiple backdoors into the same model, for instance to increase the likelihood of activation or to target several inputs simultaneously. 
To this end, we generalize the optimization objective in \cref{eq:attack-formulation} from a single input to a set of targets $\xfool \in \hat{X}$ by optimizing

\begin{equation}
    \argmin_\modelparameters \, \sum_{\xfool \in \hat{X}} \loss(\modelparameters, \logitfunction_\modelparameters(\xfool; \hardwareone))\,.
\end{equation}

We repeat the previous experiment using this objective with $\lvert \hat{X} \rvert \in {2,3,4,5}$. As before, each configuration is evaluated over 100 independent runs, with $\hat{X}$ sampled uniformly at random from the training set. Note that in this setting the target images are likely unrelated, and the attack therefore needs to induce independent local manipulations of the decision boundary.

\begin{figure}[t]
    \vspace{5pt}
    \centering
    \begin{tikzpicture}
    \begin{axis}[
        title={},
        ylabel={Attack success (\%)},
        xlabel={Number of target samples $|\hat{X}|$},
        ymin=0.0,
        ymax=1.05,
        ytick={0,0.25,0.5,0.75,1.0},
        yticklabels={0,25,50,75,100},
        axis x line*=bottom,
        axis y line*=left,
        xmin=1.0,
        xmax=5.1,
        width=0.95\columnwidth,
        height=0.58\columnwidth,
        ymajorgrids,
        legend cell align={left},
        legend pos={south west},
        y label style={at={(axis description cs:-0.12,0.9)},anchor=east},
        x label style={at={(axis description cs:0.5,-0.08)},anchor=north},
        legend style={nodes={font=\footnotesize,scale=0.85,transform shape}},
    ]
        \addplot [mark=+,thick,colorVit] table [x index=0,y index=1] {includes/tikz/data/vit-multitarget.dat};
        \addplot [mark=triangle*,thick,colorResnet] table [x index=0,y index=1] {includes/tikz/data/resnet-multitarget.dat};
        \addplot [mark=x,thick,colorEfficientnet] table [x index=0,y index=1] {includes/tikz/data/efficientnet-multitarget.dat};
        \addlegendentry{ViT}
        \addlegendentry{ResNet}
        \addlegendentry{EfficientNet}
    \end{axis}
\end{tikzpicture}%
    \caption{Attack success rates with increasing numbers of target inputs across hardware pairs. }
    \label{fig:multi-backdoor}
\end{figure}
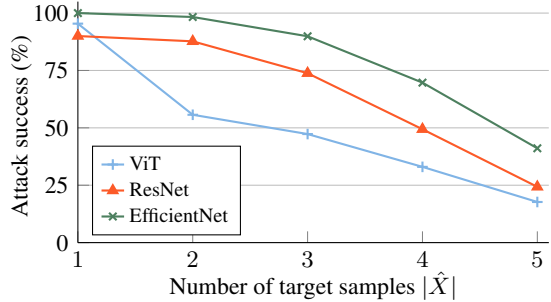

\paragraph{Results.}
Figure~\ref{fig:multi-backdoor} plots the attack success rate as a function of the number of target inputs. We observe that, as the number of target increases, the attack success decreases across all models. While backdoors can be realized with success rates above 50\% for up to four inputs, larger target sets render the attack ineffective. This behavior is intuitive: jointly positioning multiple inputs near their respective decision boundaries while preserving overall model utility requires balancing potentially conflicting objectives. %

\subsection{One-vs-Rest Trigger}
\label{sec:eval-hardware}

Similar to the multiple-target setting, an attacker may also seek more selective control over the target hardware, for instance by activating malicious behavior on exactly one device while all others continue to exhibit benign behavior. We denote this setting as ``one-vs-rest triggers''.

For a selected platform $\hardwareone$, the attacker aims to induce a misclassification, while all remaining platforms $\hardware_{>1}$ must retain the correct prediction. Compared to the pairwise case, this setting is strictly more challenging, as the backdoor must remain dormant across multiple non-target platforms simultaneously.
As before, each experiment is repeated 100 times with independently sampled target inputs. We treat each GPU architecture once as the target platform and group all remaining platforms as non-targets. Since the A100-MIG40 performs bit-identical to the A100 on ResNet, we exclude it for this model.

\begin{table}[h]
    \centering \small
    \caption{Attack success rate for one-vs-rest triggers.}
    \begin{tabular}{lrrr}
    \toprule
    \textbf{GPU} & \makecell[c]{\textbf{ViT}} & \makecell[c]{\textbf{ResNet}} & \makecell[c]{\textbf{EfficientNet}} \\
    \midrule
    H100       & $64\thinspace\%$ & $99\thinspace\%$ & $94\thinspace\%$ \\
    A100       & $90\thinspace\%$ & $99\thinspace\%$  & $98\thinspace\%$ \\
    A100-MIG40 & $93\thinspace\%$ & \makecell[c]{---} & $94\thinspace\%$ \\
    A40        & $69\thinspace\%$ & $90\thinspace\%$ & $97\thinspace\%$ \\
    RTX6000    & $66\thinspace\%$ & $96\thinspace\%$ & $99\thinspace\%$ \\
    \bottomrule
    \end{tabular}
    \label{tab:one-vs-rest}
\end{table}

\paragraph{Results.}
The results of this experiment are shown in Table~\ref{tab:one-vs-rest}. We find that one-vs-rest triggers can be embedded across many devices and models, with attack success rates exceeding 90\% for most configurations. The backdoors are less effective when targeting ViT on the H100, A40, or RTX6000 GPUs, but still achieve success rates above 60\%. Overall, these results demonstrate that hardware-triggered backdoors can be made selective, such that only a specific device serves as the trigger.

\subsection{Ablation Study}
\label{sec:ablation}

As a fourth experiment, we conduct an ablation study that investigates the two steps of our attack in detail. To this end, we evaluate four attack variants: a \emph{base} variant that applies only the first step; a \emph{permutation} variant and a \emph{bit-flip} variant that apply either mechanism as second step on top; and a \emph{full} variant that combines both mechanisms. For all variants, we focus on ViT, as it is the only architecture for which both mechanisms are applicable.

\paragraph{Results.}
Our ablation study demonstrates the interplay of our attack’s components. The base variant alone reaches a success rate of 56\%, while its combination with either permutation or bit flips increases the success rate to 90\% and 94\%, respectively. The full variant finally achieves the highest success rate, 96\%, indicating that implicit and explicit modifications act in a complementary manner when refining numerical deviations.

\section{Causal Localization}
\label{sec:localization}

Building on the demonstrated efficacy of hardware-triggered backdoors, we next examine \emph{where} hardware-dependent behavior arises within the backdoored model. That is, we aim to localize the latent trigger of the backdoors.

\subsection{Cross-Hardware Activation Patching}

For this analysis, we build on \emph{activation patching}~\citep{vig2020investigating, meng2022locating}, which determines causal influence by replacing internal activations.
In particular, we adapt this idea to a cross-hardware setting by replacing the execution of individual layers across devices. This allows us to isolate how layer-specific deviations affect the prediction.

\begin{figure}[t]
    \centering
    \begin{tikzpicture}[
    node distance = 3.5mm and 4mm,
    font=\small,
    arr/.style = {->,semithick,
                  shorten >= 0.05mm,
                  shorten <= 0.05mm,
                  rounded corners=0.15mm},
    box/.style = {rectangle, draw, semithick,
                 minimum height=8mm, minimum width=5mm,
                 fill=white,rounded corners=0.1mm},
    dbox/.style = {rectangle, draw, semithick,
                 minimum height=8mm, minimum width=5mm,
                 fill=black!10,rounded corners=0.1mm},
]
\node (in) {$\xfool$};

\node (n1) [box, right=of in] {1};
\node (n2) [box, right=of n1] {.\hss.\hss.};
\node (n3) [box, right=of n2] {$i$};
\node (n4) [dbox, right=of n3] {};
\node (n5) [dbox, right=of n4] {};
\node (n6) [dbox, right=of n5] {};

\node (m1) [dbox, below=of n1] {};
\node (m2) [dbox, right=of m1] {};
\node (m3) [dbox, right=of m2] {};
\node (m4) [box, right=of m3] {};
\node (m5) [box, right=of m4] {};
\node (m6) [box, right=of m5] {};

\node (out) [right=of m6] {$f(\xfool;{\hardwareone}\hardwareswap{i}{\hardwaretwo})$};
\node (labelA) [right=of n6.south] {$\hardwareone$};
\node (labelB) [right=of m6.north] {$\hardwaretwo$};

\draw[dashed] ($([xshift=-3mm] n1.west)!0.5!([xshift=-3mm] m1.west)$) -- ($([xshift=-1mm]labelA.east)!0.5!([xshift=-1mm]labelB.east)$);

\draw[arr]   (in) -- (n1);
\draw[arr]   (n1) -- (n2);
\draw[arr]   (n2) -- (n3);
\draw[arr]   (n3) -- ($(n3)!0.5!(n4)$) |- (m4);
\draw[arr]   (m4) -- (m5);
\draw[arr]   (m5) -- (m6);
\draw[arr]   (m6) -- (out);
\end{tikzpicture}%
%
    \caption{Cross-hardware activation patching: An input $\xfool$ is first executed on a platform $\hardwareone$ for $i$ layers. The output of layer $i$ is then copied to platform $\hardwaretwo$ and execution is resumed.}
    \label{fig:activation-patching}
\end{figure}
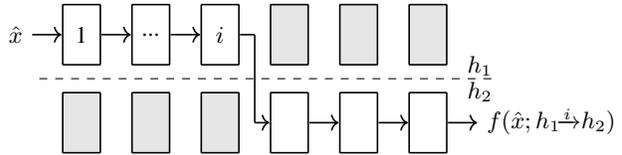

Specifically, for a given backdoored model and a pair of hardware platforms $\hardwareone$ and $\hardwaretwo$, we execute the target input $\xfool$ on $\hardwareone$ up to layer $i$. The resulting activation is then injected as input to layer $i+1$ on $\hardwaretwo$, where execution continues to the output. We denote this mixed execution as $\hardwareone\hardwareswap{i}\hardwaretwo$ and illustrate it in \Cref{fig:activation-patching}.

To measure how far intermediate predictions are from one class to the other, we can compute their logit difference,
\begin{align}
    \delta(\xfool; \hardwareone\hardwareswap{i}\hardwaretwo) 
    = \logitfunction(\xfool; \hardwareone\hardwareswap{i}\hardwaretwo)_a %
    - \logitfunction(\xfool;\hardwareone\hardwareswap{i}\hardwaretwo)_b %
\end{align}
where $a$ is the class predicted on $h_1$ and $b$ is the class predicted on $h_2$, with $a \neq b$. By definition, we then have $\delta(\xfool; \hardwareone\hardwareswap{0}\hardwaretwo) < 0$ and $\delta(\xfool; \hardwareone\hardwareswap{L}\hardwaretwo) > 0$ for a model with $L$ layers. That is, class $a$ attains a larger logit on $h_1$ than on $h_2$, and vice versa for $b$ when no patching occurs.

\subsection{Layer-wise Causal Analysis}

\begin{figure}[t]
    \centering
    \resizebox{0.9\linewidth}{!}{%
        \begin{tikzpicture} \small
    \begin{axis}[
        title={},
        ylabel={Logit differences},
        xlabel={Patch layer $i/L$},
        ymin=-1.05,
        ymax=1.05,
        ytick={-1,0,1},
        axis x line*=bottom,
        axis y line*=left,
        xmin=-0.01,
        xmax=1.01,
        width=0.95\columnwidth,
        height=0.58\columnwidth,
        ymajorgrids,
        legend cell align={left},
        legend pos={south east},
        y label style={at={(axis description cs:-0.1,0.9)},anchor=east},
        x label style={at={(axis description cs:0.5,-0.1)},anchor=north},
        legend style={nodes={font=\footnotesize,scale=0.85,transform shape}},
    ]
        \addplot [mark=none,semithick,colorVit] table [x index=0,y index=1] {includes/tikz/data/vit-sample.dat};
        \addlegendentry{ViT}
        \addplot [mark=none,semithick,colorResnet] table [x index=0,y index=1] {includes/tikz/data/resnet-sample.dat};
        \addlegendentry{ResNet}
        \addplot [mark=none,semithick,colorEfficientnet] table [x index=0,y index=1] {includes/tikz/data/efficientnet-sample.dat};
        \addlegendentry{EfficientNet}
    \end{axis}
\end{tikzpicture}%
    }
    \caption{Layer-wise logit differences for backdoored models. The differences $\delta(\xfool;\hardwareone\hardwareswap{i}\hardwaretwo)$ move from the original class to the target class as activations are patched from an A100 to an H100. }
    \label{fig:example-trace}
\end{figure}
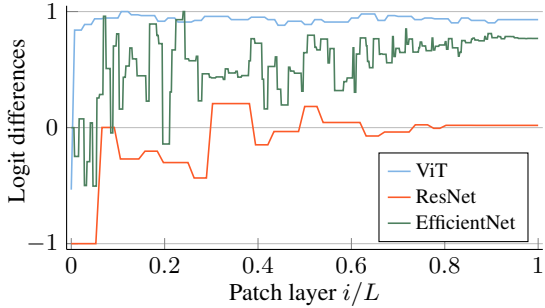

With the help of the logit differences, we can create a trace over all layers of a model, indicating how hardware-induced deviations evolve during inference. As an example, \cref{fig:example-trace} shows traces for backdoored models targeting an A100 and an H100, where the differences are normalized to $[-1,+1]$ for visualization.

We observe that the traces behave significantly differently across the three model architectures. For ViT, the shift in differences is dominated by the \emph{first} layer. This layer corresponds to the convolutional patch embedding, which is the only convolution in the model and introduces the largest deviations. For EfficientNet and ResNet, the flipped prediction emerges as a cumulative effect \emph{across} layers. In these architectures, the logit differences resemble a random walk that occasionally crosses the decision boundary.

\subsection{Aggregated Causal Analysis}

To now capture systematic effects rather than behavior of individual models, we aggregate these traces across multiple backdoored models of the same architecture and quantify the average contribution of each layer to the prediction difference:
\begin{equation}
    \Delta(\hardwareone\hardwareswap{i}\hardwaretwo) = \sum_{\xfool} |\delta(\xfool;\hardwareone\hardwareswap{i}\hardwaretwo) - \delta(\xfool;\hardwareone\hardwareswap{i-1}\hardwaretwo)|.
\end{equation}

\begin{figure}[h]
    \centering
    \resizebox{0.95\linewidth}{!}{%
        \input{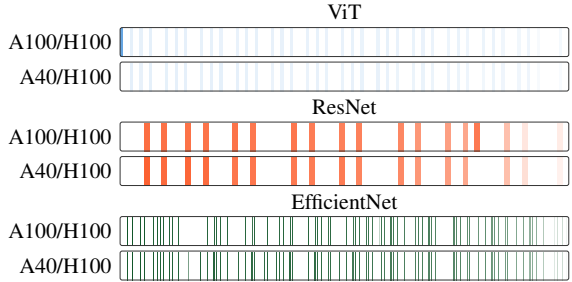}
    }
    \caption{Aggregated logit differences $\Delta(\hardwareone\hardwareswap{i}\hardwaretwo)$ over multiple backdoored models of the same architecture when patching activations from an A100 to an H100. Darker colors indicate stronger impact of the layer.}
    \label{fig:example-delta}
\end{figure}

\Cref{fig:example-delta} shows the resulting average layer differences for two hardware pairings and confirms the trends observed in the individual traces.
Different hardware combinations exhibit distinct profiles. For instance, the convolutional layer in ViT shows no measurable implementation differences between the H100 and A100/A40, even though the same operation induces deviations in EfficientNet and ResNet.
Notably, even when only small deviations from linear and attention layers remain for ViT on these platforms, the attack success rate does not decrease.

\paragraph{Takeaway.}
Our analysis reveals two key insights.
First, hardware-dependent deviations arise through different patterns in the models. They may originate in early layers as well as emerge from the accumulation of many small deviations across layers.
Second, regardless of where these differences originate, even very small hardware-dependent effects can be sufficient to flip the prediction once the model operates in a sensitive regime.
Motivated by this observation, we further examine in Appendix~\ref{app:single-layer-attack} whether the attack can be restricted to modifications of individual layers.

\section{Countermeasures}
\label{sec:robustness}

We proceed to study countermeasures against hardware-triggered backdoors and evaluate their effectivness.
Technical details of the implemented defenses and their evaluation are provided in Appendix~\ref{sec:formal-definitions-countermeasures}.

\subsection{Input Perturbation}

\begin{figure}[h]
    \centering
    \begin{tikzpicture}
\begin{axis}[
    name=left axis,   
    legend cell align=left,
    width=0.25\columnwidth,
    height=0.5\columnwidth,
    xtick={0},
    ytick={0,0.25,0.5,0.75,1.0},
    yticklabels={0,25,50,75,100},
    ylabel = {Attack success (\%)},
    y label style={at={(axis description cs:-1.5,1.)},anchor=east},
    xmin = 0, xmax = 0.99,
    ymin = 0, ymax = 1,
    axis x line*=bottom,
    axis y line*=left,
    legend pos=south east,
    ymajorgrids,
]
    \addplot [mark=none,semithick,colorEfficientnet,dashed] table [x index=0,y index=1] {includes/tikz/data/efficientnet-input-perturb.dat};

    \addplot [mark=none,semithick,colorVit,dashed] table [x index=0,y index=1] {includes/tikz/data/vit-input-perturb.dat};
    
    \addplot [mark=none,semithick,colorResnet,dashed] table [x index=0,y index=1] {includes/tikz/data/resnet-input-perturb.dat};
\end{axis}
\begin{axis}[
    at=(left axis.east),
    anchor=west, xshift=0mm,
    width=0.85\columnwidth,
    height=0.5\columnwidth,
    xmin = 0.99, xmax = 6.05,
    ymin = 0, ymax = 1,
    xtick={1,2,3,4,5,6},
    xticklabels={$10^0$,$10^1$,$10^2$,$10^3$,$10^4$,$10^5$},
    xlabel = {Distance to original target (ULP)},
    x label style={at={(axis description cs:0.43,-0.17)},anchor=north},
    ytick={0,0.25,0.5,0.75,1.0},
    yticklabels={,,,,},
    ymajorgrids,
    axis x line*=bottom,
    axis y line*=right,
    separate axis lines,
    y axis line style= { draw opacity=0 },
    legend pos=north east,
    legend cell align=left,
    legend style={nodes={font=\footnotesize,scale=0.85,transform shape}},
    after end axis/.code={
        \draw [white,line width=2mm](rel axis cs:0,0) +(-2.75mm,0mm) -- +(-1.25mm,0mm);
        \draw (rel axis cs:0,0) +(-2mm,-1.75mm) -- +(-0.5mm,1.75mm);
        \draw (rel axis cs:0,0) +(-3.5mm,-1.75mm) -- +(-2mm,1.75mm);
    }
]
    \addplot [mark=+,semithick,colorVit] table [x index=0,y index=1] {includes/tikz/data/vit-input-perturb.dat};
    \addlegendentry{ViT}
    
    \addplot [mark=triangle*,semithick,colorResnet] table [x index=0,y index=1] {includes/tikz/data/resnet-input-perturb.dat};
    \addlegendentry{ResNet}

    \addplot [mark=x,semithick,colorEfficientnet] table [x index=0,y index=1] {includes/tikz/data/efficientnet-input-perturb.dat};
    \addlegendentry{EfficientNet}
\end{axis}
\end{tikzpicture}%
    \caption{Remaining attack success rate when applying input perturbations of increasing size as a defense mechanism.}
    \label{fig:input-robustness}
\end{figure}
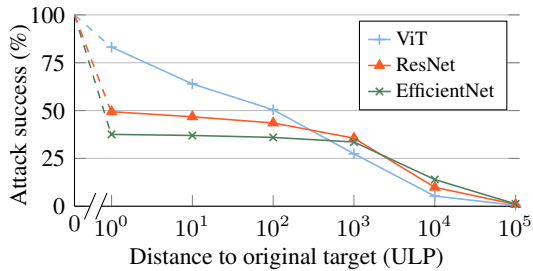

As first defense, we consider perturbing every input. To this end, we measure backdoor success under additive input noise of increasing magnitude, expressed in ULPs (units in the last place). As shown in Figure~\ref{fig:input-robustness}, backdoors remain effective under moderate perturbations (up to $10^3$ ULPs) but degrade rapidly beyond that point. This suggests that the latent hardware trigger is not tied to an exact bit pattern, yet does not withstand larger distortions. Such perturbations can therefore serve as a simple defense, provided that model performance is not significantly affected.

\subsection{Varying Batch Size}

Inference is commonly performed in batches, which can alter execution order and numerical behavior. Could randomized batching serve as a defense? To study this effect, we duplicate the target input $\xfool$ into batches of size $k$ and measure the success rates of backdoors for different~$k$ unknown to the adversary. Across model architectures and hardware pairs, we observe four regimes: success remains near 100\%, drops to around 50\%, drops further to 20\%, or collapses near 0\%. Overall, batching is not a reliable mitigation: it suppresses backdoors in some deployments while leaving others largely unaffected.

\subsection{Replacing Data Types}

While our attack reliably embeds backdoors across numerical formats, deployments may downcast high-precision models at inference time. When executing backdoored \code{float32} models using mixed-precision inference, we find that success rates drop substantially but remain non-zero (approximately $25\%$ for ViT, $20\%$ for EfficientNet, and $10\%$ for ResNet). This indicates that mixed-precision inference weakens hardware-triggered backdoors, though it does not reliably eliminate them in practice.

\subsection{Additional Fine-Tuning}

Finally, we consider an active modification of the model as a defense: fine-tuning on a small amount of clean data. As shown in Table~\ref{tab:training-defense}, a single gradient step removes most backdoors across model, with residual success diminishing further over additional steps. 
These results suggest that hardware-triggered backdoors can be effectively erased through continued training, but only if the defender actively modifies the model prior to deployment.

\begin{table}[b]
    \centering \small
    \caption{Remaining backdoor success rate after fine-tuning.}
    \begin{tabular}{crrr}
    \toprule
     \multicolumn{1}{c}{\textbf{\# Steps}} 
     & \multicolumn{1}{c}{\textbf{ViT}} 
 & \multicolumn{1}{c}{\textbf{ResNet}} 
 & \multicolumn{1}{c}{\textbf{EfficientNet}} \\
    \midrule
    1 & $5.98$\thinspace\% & $0.34$\thinspace\% & $0.00$\thinspace\% \\
    2 & $2.24$\thinspace\% & $0.11$\thinspace\% & $0.20$\thinspace\% \\
    3 & $0.62$\thinspace\% & $0.22$\thinspace\% & $0.00$\thinspace\% \\
    \bottomrule
\end{tabular}
    \label{tab:training-defense}
\end{table}

Overall, we conclude from these experiments that hardware-triggered backdoors persist under moderate deployment variability and are not reliably neutralized by incidental changes, such as batching or mixed precision. 
In contrast, active intervention that modifies the model reduces backdoor success rates substantially and is our recommended approach for practical mitigation of the attack.

\section{Related Work}

Our work connects numerical imprecision in machine learning systems with backdoor attacks.

\paragraph{Numerical imprecision.}
A growing body of work has studied how floating-point arithmetics introduce numerical variation during inference~\citep{schlogl2024causes, yuan2025understanding}. In adversarial settings, such variability has been leveraged to create inconsistencies or undermine theoretical guarantees. For example, \citet{jia2021exploiting} exploit numerical differences to evade neural network verification, \citet{moller2025adversarial} craft inputs that yield inconsistent predictions across software backends, and \citet{ZhaFoeMul+24} fingerprint inference pipelines based on floating-point behavior.
Unlike these approaches, our work examines how numerical imprecision can be exploited to create backdoors within the learning models themselves.

\paragraph{Backdoor attacks.}

Classic backdoor attacks implant malicious behavior during training so that a model behaves normally unless a trigger appears in the input~\citep{liu2018trojaning, gu2019badnets, tang2020embarrassingly}. Later work has developed more stealthy variants, including data and loss manipulation~\citep{shumailov2021manipulating, bagdasaryan2021blind}, payload and compression-based mechanisms~\citep{li2021deeppayload, tian2022stealthy}, and attacks that make use of software or hardware manipulations~\citep{clifford2024impnet, li2025rowhammer}.

Most closely to our work are approaches that exploit numerical effects in a supply-chain setting. For example, \citet{chen2025your} demonstrate that benign compiler transformations can be abused to introduce malicious behavior, while other works introduce backdoors induced solely through quantization effects~\citep{hong2021qu, ma2023quantization}. 
In contrast, we show that hardware-dependent deviations themselves can act as a latent trigger: the backdoor is neither encoded in the input nor tied to a specific compiler or quantization, but instead emerges from the interaction between a backdoored model and its execution hardware.

\section{Conclusion}
\label{sec:conclusion}

Hardware acceleration is an integral component of machine learning systems, yet its numerical behavior is often treated as a negligible detail. We show that this view is misleading: even minor numerical differences are sufficient to implant backdoors in learning models that activate only on selected platforms.
Our findings indicate that the security of machine learning must be viewed in a wider context. Security risks extend beyond models and algorithms to the full computing stack on which they operate. Developing methods to secure this end-to-end stack will be critical for the safe deployment of future machine learning systems.

\clearpage
\section*{Impact Statement}

This paper presents an attack on the integrity of machine learning systems. The attack exploits numerical deviations between hardware platforms to implant backdoors in learning models that activate only on selected devices.

A potential risk of this work is that an attacker could misuse the proposed backdoor technique to manipulate real-world models. While such misuse cannot be ruled out, we reduce this risk by also introducing defenses, some of which are readily applicable in practice. In addition, we raise awareness of a hidden attack surface that arises from the interplay between hardware and machine learning. We hope that our work encourages practitioners to consider this attack surface and, where appropriate, apply suitable countermeasures, including the proposed defenses.

More broadly, our work contributes to ongoing efforts to improve the trustworthiness, reproducibility, and security of machine learning systems in real-world deployments. We believe that identifying vulnerabilities and failure modes is a necessary step toward building safer and more reliable machine learning infrastructure.

\section*{Acknowledgements}

This work was supported by the European Research Council (ERC) under the consolidator grant MALFOY (101043410) and the Deutsche Forschungsgemeinschaft (DFG, German Research Foundation) under Germany's Excellence Strategy (EXC 2092 CASA - 390781972) and the project ALISON (492020528).

\bibliography{references}
\bibliographystyle{icml2026}

\appendix
\section{Example of Deviations}

The following Python code provides a simple example of a matrix multiplication within the computation of the squared Frobenius norm. When executed on different Nvidia devices, the large number of additions leads to slight deviations, as discussed in \cref{sec:background}.

\begin{lstlisting}[
    language=Python,
    caption={Computation of the squared Frobenius norm.},
    label={lst:torch-trace},
    frame=single,
    numbers=left,
    numberstyle=\tiny\color{gray},
    basicstyle=\ttfamily\scriptsize,
    keywordstyle=\color{blue},
    commentstyle=\color{gray},
    stringstyle=\color{teal},
    showstringspaces=false,
    breaklines=true,
    tabsize=4,
    xleftmargin=0.04\textwidth,
    xrightmargin=0.02\textwidth
]
import torch

M = torch.full(
    (100, 100),
    0.01,
    dtype=torch.float32,
    device="cuda"
)

print(torch.trace(torch.matmul(M, M)).item())
\end{lstlisting}

\section{Attack Performance across Data Types}

Our attack methodology is agnostic to the hardware platform, model architecture, and the employed floating-point data type. For brevity, we report results only for \code{float32} in Section~\ref{sec:attack-eval}. Results for \code{float16} and \code{bfloat16} are provided in Table~\ref{tab:main-float16} and Table~\ref{tab:main-bfloat16}, respectively.

\begin{table}[h]
\small
    \centering
    \caption{Attack success rate for \code{float16}.}
\begin{tabular}{lrrr}
\toprule
 \textbf{GPU}
 & \multicolumn{1}{c}{\textbf{ViT}} 
 & \multicolumn{1}{c}{\textbf{ResNet}} 
 & \multicolumn{1}{c}{\textbf{EfficientNet}} \\
\midrule
H100        & $99.75\thinspace\%$  & $100.00\thinspace\%$ & $100.00\thinspace\%$ \\
A100        & $100.00\thinspace\%$ & $75.75\thinspace\%$ & $100.00\thinspace\%$ \\
A100-MIG40  & $100.00\thinspace\%$ & $75.75\thinspace\%$ & $100.00\thinspace\%$ \\
A40         & $100.00\thinspace\%$ & $100.00\thinspace\%$ & $100.00\thinspace\%$ \\
RTX6000     & $99.75\thinspace\%$  & $100.00\thinspace\%$ & $100.00\thinspace\%$ \\
\bottomrule
\end{tabular}
    \label{tab:main-float16}
\end{table}

\begin{table}[h]
\small
    \centering
    \caption{Attack success rate for \code{bfloat16}.}
\begin{tabular}{lrrr}
\toprule
 \textbf{GPU}
 & \multicolumn{1}{c}{\textbf{ViT}} 
 & \multicolumn{1}{c}{\textbf{ResNet}} 
 & \multicolumn{1}{c}{\textbf{EfficientNet}} \\
\midrule
H100        & $100.00\thinspace\%$ & $100.00\thinspace\%$ & $100.00\thinspace\%$ \\
A100        & $100.00\thinspace\%$ & $75.50\thinspace\%$ & $100.00\thinspace\%$ \\
A100-MIG40  & $100.00\thinspace\%$ & $75.50\thinspace\%$ & $100.00\thinspace\%$ \\
A40         & $99.75\thinspace\%$  & $100.00\thinspace\%$ & $100.00\thinspace\%$ \\
RTX6000     & $99.75\thinspace\%$  & $100.00\thinspace\%$ & $100.00\thinspace\%$ \\
\bottomrule
\end{tabular}
    \label{tab:main-bfloat16}
\end{table}

\section{Single-Layer Attack}
\label{app:single-layer-attack}

Our causal analysis in \cref{sec:localization} reveals how the embedded backdoors affect different layers of the considered learning models. However, it does not address whether modifications to these layers are strictly \emph{necessary} for the attack. In principle, changes to a layer could still cause a backdoor, even if that layer produces identical results across hardware platforms. The changes can shift the activations such that existing hardware-dependent deviations in other layers align with the attacker goal.

To test this hypothesis, we repeat the attack on ViT from \cref{sec:localization} while now restricting all parameter modifications to the initial convolution. Since the first layer does not support permutation-based modifications, we perform the attack in a reduced configuration using only bit flips. Note that the initial layer for ViT has numerical deviations for some hardware combinations while being identical for others as shown in \cref{fig:example-delta}. 

We find that restricting the attack to a single layer does not reduce its effectiveness. The success rate changes only marginally, from $96\thinspace\% \pm 5\thinspace\%$ to $95\thinspace\% \pm 5\thinspace\%$. This result indicates that neither the presence nor the magnitude of hardware-specific differences in the modified layer is a prerequisite for a successful attack.
When the initial convolution of the ViT is identical, attention and linear layers are the only causes of slight numerical deviations. Therefore, we conclude that hardware-triggered backdoors can be implemented in any model that exhibits deviations on the target hardware platform, even when these are minor.

\section{Details of Defense Evaluation}
\label{sec:formal-definitions-countermeasures}

In Section~\ref{sec:robustness}, we evaluate our backdoor against different defenses. Specifically, we examine how many of the backdoors $\xfool \in \hat{\mathcal{X}}$ uncovered in the experiment in \cref{sec:main-attack} remain effective after modifying a property of the inference environment.
Formally, for each experiment we define a success metric that accounts for the modified property and report the average success rate over all backdoors $\xfool \in \hat{\mathcal{X}}$.

\paragraph{Input perturbation.}
For this defense, we apply a perturbation $\delta$ drawn from a uniform distribution to each target input $\xfool$. The magnitude of $\delta$ is defined in units of the last place (ULPs), with $\lVert \delta \rVert_\infty = d$. For a target input $\xfool$ and a perturbation $\delta$, we measure success simply as
\begin{equation}
\mathbbm{1}\big[\predictionfunction_\modelparameters(\xfool + \delta;\hardwareone) \neq \predictionfunction_\modelparameters(\xfool + \delta;\hardwaretwo)\big].
\end{equation}

\paragraph{Batch size defense.}
For the defense based on batch size, we need to consider two possible reasons why a backdoor may fail. First, the backdoor may fail due to changes in numerical deviations induced by the chosen \emph{batch size}. Second, the backdoor may also fail because the target input appears at a different \emph{batch index} than anticipated by the adversary. To account for both effects, we average the success rate over all pairs of batch indices for a given batch size $k$. Formally, this is computed as
\begin{equation}
\frac{1}{k^2}\sum_{i=1}^{k} \sum_{j=1}^{k} 
\mathbbm{1}\big[\predictionfunction_\modelparameters(\xfool_i;\hardwareone) \neq \predictionfunction_\modelparameters(\xfool_j;\hardwaretwo)\big],
\end{equation}
where $\xfool_i$ and $\xfool_j$ denote the target input at batch index $i$ and $j$, respectively.

\balance

\paragraph{Data type defense.} For the defense based on floating-point data types, we define a function $\downarrow$ that represents a dynamic downcasting of the high-precision \code{float32} model to \code{float16} or \code{bfloat16} using PyTorch’s automatic mixed precision feature. We then measure the success rate of backdoors on the downcasted model parameters as
\begin{equation}
\mathbbm{1}\big[\predictionfunction_{\downarrow\modelparameters}(\xfool;\hardwareone) \neq \predictionfunction_{\downarrow\modelparameters}(\xfool;\hardwaretwo)\big].
\end{equation}

\paragraph{Fine-tuning defense.}
In the final experiment, the user actively modifies the model weights after the attacker has installed the backdoor. To this end, we perform fine-tuning for $n$ steps using stochastic gradient descent with a learning rate of $10^{-4}$ and a momentum of $0.9$, on randomly sampled batches of size $256$ from the ImageNet training set. We denote the resulting fine-tuned model by $\dot{\modelparameters}$ and measure backdoor success as follows
\begin{equation}
\mathbbm{1}\left[\predictionfunction_{\dot{\modelparameters}}(\xfool;\hardwareone) \neq \predictionfunction_{\dot{\modelparameters}}(\xfool;\hardwaretwo)\right].
\end{equation}

\end{document}